\documentclass[letterpaper]{article}
\usepackage{aaai}
\usepackage{times}
\usepackage{helvet}
\usepackage{courier}
\usepackage{amsmath,amssymb,amsfonts}
\usepackage{graphicx}
\usepackage{textcomp}
\usepackage[table]{xcolor}
\usepackage{booktabs}
\usepackage{multirow}
\usepackage{array}
\usepackage{pifont}
\nocopyright
\title{FedLAB: Traceable Semantic Codebooks for Federated Multimodal Graph Foundation Learning}
\author{Zekai Chen*, Kairui Yang*, Xuaner Chen, Xunkai Li, Xun Wu, Rong-Hua Li, Guoren Wang\\
Beijing Institute of Technology, Beijing, China}

\begin{document}
\maketitle

\begin{abstract}
Multimodal graph foundation models aim to learn reusable knowledge from graphs enriched with text, images, attributes, and relational topology, thereby supporting diverse graph-centric and modality-centric tasks. In practice, however, such multimodal graphs are often distributed across decentralized clients, where raw contents and local structures cannot be centrally shared due to privacy constraints. This motivates federated multimodal graph foundation learning, which requires not only transferable representation learning but also intrinsic semantic traceability under strict data isolation. Existing methods usually exchange or store knowledge through parameters, prototypes, embeddings, or compact codebooks, which support optimization and transfer but do not explicitly expose how modality evidence, node semantics, and topology context jointly support predictions. To bridge this gap, we propose FedLAB, a traceable semantic codebook framework that organizes multimodal graph knowledge into typed hierarchical codebooks for modality evidence, node semantics, and topology context. FedLAB further refines these trace units through federated semantic barycenter pre-training while keeping raw multimodal contents and graph structures local. Extensive experiments on 10 benchmarks and 6 downstream tasks show that FedLAB improves over state-of-the-art baselines by up to 7.53\%, while preserving a native semantic trace interface.
\end{abstract}

\section{Introduction}

Multimodal graph foundation models (MM-GFMs), as multimodal extensions of graph foundation models (GFMs), aim to learn reusable knowledge from graphs enriched with text, images, attributes, and relational topology, thereby supporting diverse graph-centric and modality-centric tasks~\cite{graphmae,graphclip,unigraph2}. In practice, large-scale multimodal graphs are often distributed across decentralized clients, where nodes, edges, attributes, labels, and multimodal contents cannot be centrally aggregated. Federated graph learning (FGL), especially multimodal FGL (MM-FGL), provides a practical learning paradigm for this setting~\cite{openfgl,mmopenfgl}. This motivates federated multimodal graph foundation learning (Fed-MMGFM), whose goal is not merely task-specific federated optimization, but learning a reusable multimodal graph backbone that generalizes across clients, modalities, and downstream tasks under strict data isolation.

For MM-GFMs, semantic traceability refers to the ability of a foundation backbone to expose an inspectable support path from learned knowledge to task predictions, rather than only producing opaque outputs. This property is particularly important in Fed-MMGFM, where the same backbone is reused across heterogeneous clients and shifting data distributions; without traceable support, erroneous transfer, modality bias, and shortcut-driven predictions are difficult to diagnose. However, existing federated and foundation-style graph learning methods typically exchange knowledge through parameters, gradients, prototypes, embeddings, or compact codebooks~\cite{chen2026stage,chen2026prism,fedbook}. Although effective for optimization and transfer, these mechanisms do not explicitly preserve how modality signals, node semantics, and topology jointly support predictions. Post-hoc graph explainers can identify influential features or subgraphs for a trained model~\cite{gnnexplainer,pgexplainer}, but they remain instance-specific and are not learned as reusable trace interfaces during federated pre-training. Therefore, Fed-MMGFM calls for an intrinsic trace interface jointly learned with the backbone and consistently maintained across clients.

\begin{figure}[!t]
    \centering
    \includegraphics[width=\linewidth]{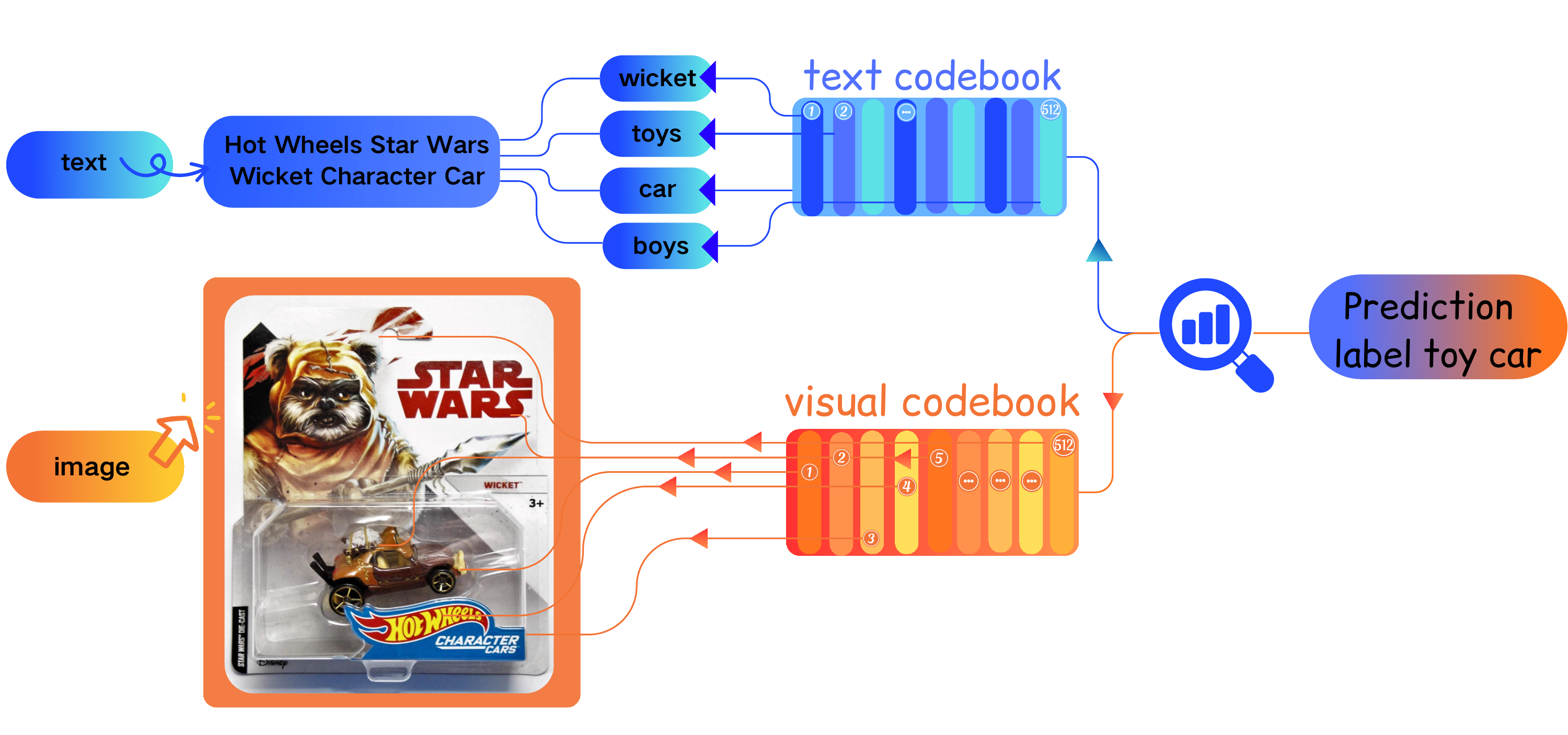}
    \caption{A traceable modality-evidence interface. FedLAB decomposes a
    multimodal prediction into text-side and image-side evidence codes rather
    than only producing a fused representation.}
    \label{fig:intro_modality_trace}
\end{figure}

We define this missing capability as the \emph{semantic traceability gap} in Fed-MMGFM. In this setting, a traceable foundation backbone should expose prediction support through three complementary interfaces. \ding{182} \textbf{Modality-Evidence Traceability}: which modality-specific signals are responsible for a prediction. \ding{183} \textbf{Node-Semantic Traceability}: which reusable node-level semantic units are consistently activated across heterogeneous clients. \ding{184} \textbf{Topology-Context Traceability}: how local structural contexts modulate or refine node semantics. Therefore, the central challenge is not only to learn a shared federated multimodal representation, but to endow it with a unified mechanism that makes modality evidence, node semantics, and topology context explicitly traceable and reusable.

To address this problem, we propose \textbf{FedLAB}, a traceable semantic codebook framework for Fed-MMGFM. FedLAB instantiates the three trace interfaces with typed hierarchical codebooks: modality evidence codebooks capture modality-specific support signals, a node semantic codebook abstracts reusable node-level concepts, and a topology context codebook models structure-induced semantic variations. For each prediction, FedLAB outputs a semantic trace path composed of evidence codes, a node semantic code, a topology context code, and contribution scores. These trace units are refined across clients through federated semantic barycenter pre-training, where clients keep raw multimodal contents and graph structures local while only sharing model updates and aggregated posterior code statistics.

\textbf{Our Contributions.} \textbf{\textit{\underline{(1) New Challenge.}}} We identify the semantic traceability gap in Fed-MMGFM, highlighting the lack of intrinsic, training-time mechanisms to trace modality evidence, node semantics, and topology context behind predictions. \underline{\textit{\textbf{(2) New Framework.}}} We propose FedLAB, a traceable semantic codebook framework that organizes multimodal graph knowledge into modality evidence, node semantic, and topology context codes, and refines them via federated semantic barycenter pre-training without sharing raw multimodal contents or local graph structures. \underline{\textit{\textbf{(3) SOTA Performance.}}} Across 10 multimodal graph datasets and 6 graph-centric and modality-centric tasks, FedLAB consistently outperforms state-of-the-art baselines by 4.38\% on average while preserving a native semantic trace interface.

\begin{figure}[t]
    \centering
    \includegraphics[width=\linewidth]{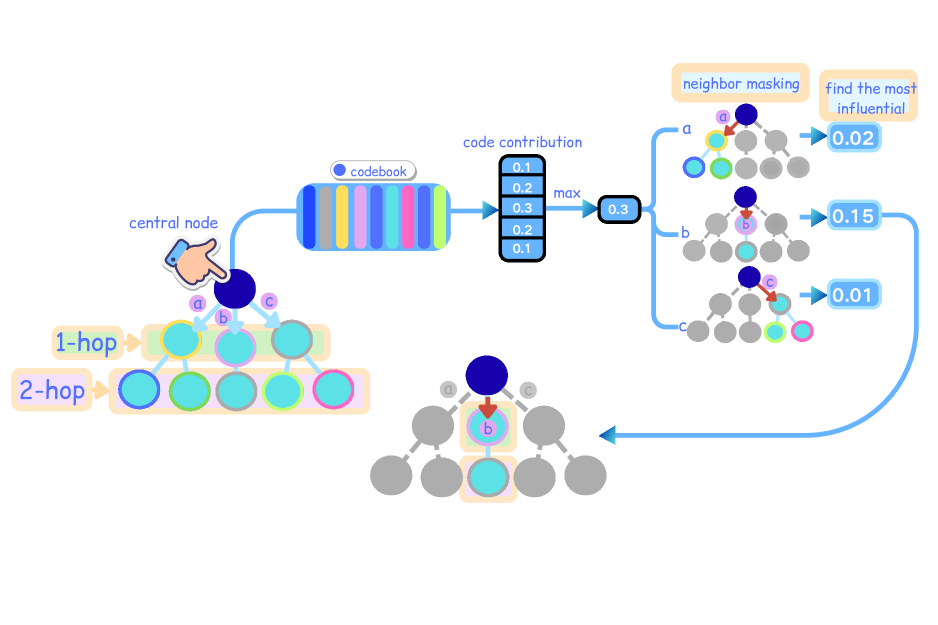}
    \caption{A traceable topology-context interface. FedLAB maps noisy graph
    neighborhoods to topology context codes and exposes the structural trace
    behind the prediction.}
    \label{fig:intro_topology_trace}
\end{figure}

\section{Preliminaries and Problem Formalization}
\label{sec:preliminaries}

\textbf{Federated Traceable MMGFM Objective.}
Following the federated multimodal graph setting in MM-OpenFGL~\cite{mmopenfgl}, we consider a server and $K$ clients, where each client $k$ owns a private multimodal graph $G_k=(V_k,E_k,\{X_k^r\}_{r\in\mathcal{M}},Y_k)$ and a modality-availability mask $M_k$. Raw multimodal contents, labels, and graph structures remain local, while the server coordinates collaborative training.

FedLAB aims to learn a federated multimodal graph foundation model that returns both a task prediction and an intrinsic semantic trace. The trace interface is instantiated by typed codebooks
\begin{equation}
\mathcal{B}
=
\left(
\{\mathcal{B}^{r}\}_{r\in\mathcal{M}},
\mathcal{B}^{s},
\mathcal{B}^{t}
\right),
\label{eq:typed_codebooks}
\end{equation}
which correspond to modality evidence, node semantics, and topology context, respectively. For node $v$ on client $k$, let $\mathcal{M}_{k,v}=\{r\in\mathcal{M}\mid M_k(v,r)=1\}$ denote its available modalities. FedLAB predicts $\hat{y}_{k,v}$ and produces
\begin{equation}
\mathcal{T}_{k}(v)
=
\left(
\{c_{k,v}^{r}\}_{r\in\mathcal{M}_{k,v}},
c_{k,v}^{s},
c_{k,v}^{t},
\Pi_{k,v}
\right),
\label{eq:trace_output}
\end{equation}
where $c_{k,v}^{r}$, $c_{k,v}^{s}$, and $c_{k,v}^{t}$ are modality-evidence, node-semantic, and topology-context trace units, while $\Pi_{k,v}$ records their contribution scores. The objective is to improve downstream utility while making these trace units reusable across clients, with only model updates and aggregated posterior code statistics shared with the server.

\section{Related Work}
\label{sec:related_work}

\textbf{Multimodal and Federated Graph Foundation Learning.}
GFMs shift graph learning from task-specific architectures toward reusable backbones learned through pre-training and adaptation~\cite{graphmae}. For multimodal graphs, recent studies further integrate relational topology with text, vision, or attributes through language-graph alignment, unified embedding spaces, or discrete graph tokenization~\cite{graphclip,unigraph2,wang2025learning}. In federated settings, MM-FGL and federated GFM methods learn transferable knowledge under data isolation through alignment signals, prototypes, anchors, or global codebooks~\cite{mmopenfgl,chen2026stage,chen2026prism,fedgfm,fedbook}. These methods improve optimization and transfer, but their shared knowledge is mainly represented as hidden states, embeddings, prototypes, or task-agnostic tokens, rather than as reusable semantic trace interfaces.

\textbf{Graph Explainability and Semantic Traceability.}
Post-hoc graph explainers identify influential nodes, subgraphs, or features for trained GNNs~\cite{gnnexplainer,pgexplainer}. However, they are usually applied after training and provide instance-level explanations instead of reusable semantic units shared across clients. FedLAB targets a different goal in Fed-MMGFM: it learns typed trace units during federated multimodal foundation pre-training, so that modality evidence, node semantics, and topology context become inspectable and reusable parts of the backbone itself.

\section{Methodology}
\label{sec:method}

\subsection{Overview}
\label{sec:method_overview}

\begin{figure*}[t]
    \centering
    \includegraphics[width=0.998\textwidth]{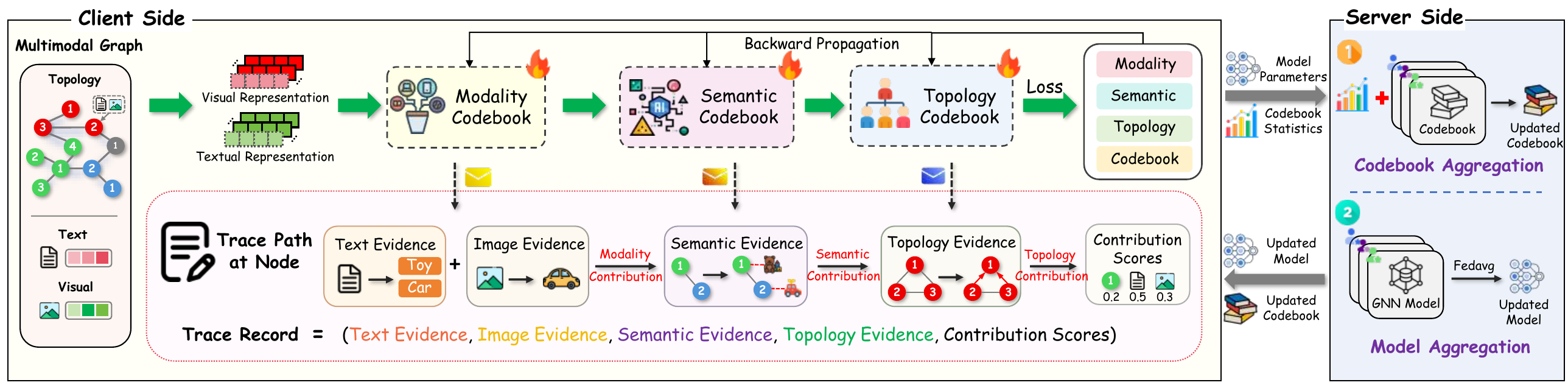}
    \caption{Framework overview of FedLAB. The server maintains a shared backbone
and typed trace codebooks, while clients encode private multimodal graphs into
hierarchical trace paths. Model updates and aggregated posterior code statistics
are uploaded to refine the backbone and shared trace vocabulary without exposing
raw data.}
    \label{fig:framework}
\end{figure*}

We present FedLAB, a traceable semantic codebook framework for Fed-MMGFM. As shown in Fig.~\ref{fig:framework}, FedLAB learns a shared backbone from private client-side multimodal graphs and returns both task representations and semantic trace paths. It replaces monolithic fused embeddings with typed semantic references for modality evidence, node semantics, and topology context. Clients generate hierarchical traces locally, while the server refines the backbone and trace vocabulary using model updates and aggregated posterior code statistics without exposing raw data.

\subsection{Shared Trace Vocabulary}
\label{sec:semantic_reference}

\textbf{Motivation.}
In Fed-MMGFM, trace units must be locally faithful while remaining comparable across clients. However, heterogeneous modality distributions and graph structures make client-local traces difficult to align, and raw contents or topology cannot be shared. FedLAB therefore introduces a shared trace vocabulary as a common semantic reference space.

\textbf{Typed trace operator.}
FedLAB represents the shared vocabulary as typed codebooks:
\begin{equation}
\mathcal{B}
=
\left(
{\mathcal{B}^{r}}*{r\in\mathcal{M}},
\mathcal{B}^{s},
\mathcal{B}^{t}
\right),
\quad
\mathcal{B}^{a}
=
{b_1^a,\ldots,b*{L_a}^a},
\label{eq:method_reference}
\end{equation}
where $a\in\mathcal{M}\cup{s,t}$. The three codebook types correspond to modality-evidence, node-semantic, and topology-context trace units, respectively. For a type-$a$ representation $h$, FedLAB first computes a posterior code assignment:
\begin{equation}
q^a(h)
=
\arg\min_{q\in\Delta^{L_a}}
\langle q,d^a(h)\rangle
+
\tau_a\Omega(q),
\label{eq:method_trace_posterior}
\end{equation}
where $d_i^a(h)=|h-b_i^a|*2^2$, $\Omega(q)=\sum_i q_i\log q_i$, and $\tau_a$ controls assignment smoothness. Based on this posterior, FedLAB first obtains the relaxed trace representation and selected index:
\begin{equation}
\begin{alignedat}{2}
\bar{z}^{a}(h) &={} \sum_{i=1}^{L_a} q_i^a(h)b_i^a, &
\ell^a(h) &={} \arg\max_i q_i^a(h).
\end{alignedat}
\label{eq:method_soft_trace}
\end{equation}
It then produces the differentiable trace representation and selected trace unit:
\begin{equation}
\begin{aligned}
z^a(h)
&={}
\mathrm{sg}\left[
b_{\ell^a(h)}^a-\bar{z}^{a}(h)
\right]
+
\bar{z}^{a}(h),\
c^a(h)
&={}
b_{\ell^a(h)}^a .
\end{aligned}
\label{eq:method_trace_readout}
\end{equation}
Here $q^a(h)$ records posterior code usage, $z^a(h)$ is used for differentiable optimization, $c^a(h)$ is the selected semantic trace unit, and $\mathrm{sg}[\cdot]$ denotes stop-gradient~\cite{van2017neural}. For simplicity, we denote the whole typed readout process as
\begin{equation}
\mathfrak{Q}_a(h;\mathcal{B}^{a})
=
(z^a(h),c^a(h),q^a(h)).
\label{eq:method_trace_operator}
\end{equation}
At each communication round, the server broadcasts the current backbone and typed codebooks to selected clients, enabling local trace encoding in a shared semantic coordinate system while raw client data remain private.

\subsection{Hierarchical Trace Encoding}
\label{sec:trace_encoding}

\textbf{Motivation.}
The shared trace vocabulary provides common semantic references, but it does not specify how prediction support is formed on each private client graph. In multimodal graphs, observed modalities first provide node-specific support, these signals are composed into reusable node semantics, and local topology further contextualizes the semantic state. If compressed into a single fused representation, these roles become entangled and difficult to inspect. FedLAB therefore performs hierarchical trace encoding, turning local forward computation into an explicit path from modality evidence to node semantics and topology context.

\textbf{Evidence-semantic routing.}
For node $v$ on client $k$, let $\mathbf{m}_{k,v}$ denote the modality-availability mask induced by $M_k(v,r)$, and let $\mathcal{M}_{k,v}$ be the set of observed modalities. For each observed modality, FedLAB projects the raw feature into the shared trace space and retrieves the corresponding modality-evidence trace unit:
\begin{equation}
\begin{gathered}
    h_{k,v}^{r}
    =
    p_r(f_r(x_{k,v}^{r})),
    \quad r\in\mathcal{M}_{k,v},\\
    (z_{k,v}^{r},c_{k,v}^{r},q_{k,v}^{r})
    =
    \mathfrak{Q}_{r}
    (h_{k,v}^{r};\mathcal{B}^{r}).
\end{gathered}
\label{eq:method_modality_code}
\end{equation}
Here $\mathfrak{Q}_{r}(\cdot)$ is the typed trace readout defined in Sec.~\ref{sec:semantic_reference}; $c_{k,v}^{r}$ records modality-side support, while $q_{k,v}^{r}$ is used for optimization and federated codebook refinement.

FedLAB then composes modality evidence into a node-semantic trace. It computes modality support scores and normalizes them over observed modalities:
\begin{equation}
\begin{gathered}
    s_{k,v}^{r}
    =
    \mathbf{w}_{m}^{\top}
    \sigma
    \left(
    W_m[z_{k,v}^{r};h_{k,v}^{r}]
    \right),\\
    \rho_{k,v}
    =
    \operatorname{MSoftmax}
    \left(
    (s_{k,v}^{r})_{r\in\mathcal{M}}/\tau_m,
    \mathbf{m}_{k,v}
    \right).
\end{gathered}
\label{eq:method_semantic_routing}
\end{equation}
where $\operatorname{MSoftmax}$ denotes masked softmax that excludes missing modalities. The routed modality evidence is summarized and then read from the node semantic codebook:
\begin{equation}
\begin{gathered}
    u_{k,v}^{r}
    =
    \varphi_m([z_{k,v}^{r};h_{k,v}^{r}]),
    \bar{h}_{k,v}^{m}
    =
    \sum_{r\in\mathcal{M}_{k,v}}
    \rho_{k,v}^{r}u_{k,v}^{r},\\
    (z_{k,v}^{s},c_{k,v}^{s},q_{k,v}^{s})
    =
    \mathfrak{Q}_{s}
    (g_s(\bar{h}_{k,v}^{m});\mathcal{B}^{s}).
\end{gathered}
\label{eq:method_semantic_code}
\end{equation}
Therefore, the node-semantic code is the reusable semantic unit activated after modality evidence has been composed, rather than a direct opaque fused embedding.

\textbf{Topology-context routing.}
FedLAB further contextualizes the node-semantic trace through local topology. Let $\chi_{k,v}$ denote a lightweight structural descriptor of node $v$, such as degree or local neighborhood statistics. For each neighbor $u\in\mathcal{N}_{k}(v)$, FedLAB scores its topology contribution by
\begin{equation}
\begin{gathered}
    \eta_{k,v,u}
    =
    [z_{k,v}^{s};z_{k,u}^{s};\chi_{k,v};\chi_{k,u}],
    e_{k,v,u}
    =
    \mathbf{w}_{t}^{\top}
    \sigma(W_t\eta_{k,v,u}),\\
    \beta_{k,v}
    =
    \operatorname{Softmax}
    \left(
    (e_{k,v,u})_{u\in\mathcal{N}_{k}(v)}/\tau_t
    \right).
\end{gathered}
\label{eq:method_top_routing}
\end{equation}
Here $\beta_{k,v,u}$ denotes the contribution of neighbor $u$ to the topology-context trace. FedLAB summarizes topology-supported semantics and reads the topology-context trace by
\begin{equation}
\begin{gathered}
    \bar{z}_{k,v}^{N}
    =
    \sum_{u\in\mathcal{N}_{k}(v)}
    \beta_{k,v,u}z_{k,u}^{s},
    \bar{h}_{k,v}^{t}
    =
    \varphi_t
    \left(
    z_{k,v}^{s},
    \bar{z}_{k,v}^{N},
    \chi_{k,v}
    \right),\\
    (z_{k,v}^{t},c_{k,v}^{t},q_{k,v}^{t})
    =
    \mathfrak{Q}_{t}
    (g_t(\bar{h}_{k,v}^{t});\mathcal{B}^{t}).
\end{gathered}
\label{eq:method_topology}
\end{equation}
Thus, $\beta_{k,v}$ exposes which neighbors support the topology-context trace, while $c_{k,v}^{t}$ records the selected structure-induced semantic context.

\textbf{Trace path readout.}
Let $\mathbf{c}_{k,v}^{m}=\{c_{k,v}^{r}\mid r\in\mathcal{M}_{k,v}\}$ denote the activated modality-evidence units. FedLAB returns the prediction together with a compact semantic trace path:
\begin{equation}
\begin{gathered}
    \Pi_{k,v}
    =
    (\rho_{k,v},\beta_{k,v}),
    \mathcal{T}_{k}(v)
    =
    \left(
    \mathbf{c}_{k,v}^{m},
    c_{k,v}^{s},
    c_{k,v}^{t},
    \Pi_{k,v}
    \right).
\end{gathered}
\label{eq:method_trace}
\end{equation}
Since $\rho_{k,v}$ and $\beta_{k,v}$ directly participate in modality composition and topology routing, the trace path is part of the prediction process and connects each output to modality evidence, node semantics, and topology context.

\subsection{Semantic Barycenter Pretraining}
\label{sec:fed_pretraining}

\textbf{Motivation.}
Hierarchical trace encoding makes each client's prediction process inspectable, but its trace units may remain client-specific due to heterogeneous modalities and graph structures. For foundation learning, these local traces must be aligned into reusable cross-client semantic references. Since raw contents and topology cannot be shared, FedLAB performs semantic barycenter pretraining: clients preserve trace faithfulness locally, while the server refines the shared trace vocabulary using aggregated posterior code statistics.

\textbf{Local trace preservation.}
For modality $r$ of node $v$, FedLAB denotes its hierarchical trace state as
$\xi_{k,v}^{r}=[z_{k,v}^{r};z_{k,v}^{s};z_{k,v}^{t}]$. The local training objective is
\begin{equation}
\begin{aligned}
\mathcal{L}_{\mathrm{loc}}^{k}
=
\mathcal{L}_{\mathrm{task}}^{k}
+
\lambda_{\mathrm{tr}}\mathcal{L}_{\mathrm{trace}}^{k}
+
\lambda_{\mathrm{cb}}\mathcal{L}_{\mathrm{cb}}^{k}.
\end{aligned}
\label{eq:method_local_loss}
\end{equation}
Here $\mathcal{L}_{\mathrm{task}}^{k}$ denotes the supervised or self-supervised task objective, while $\mathcal{L}_{\mathrm{trace}}^{k}$ preserves the three trace interfaces:
\begin{equation}
\mathcal{L}_{\mathrm{trace}}^{k}
=
\mathcal{L}_{\mathrm{mod}}^{k}
+
\lambda_{\mathrm{sem}}\mathcal{L}_{\mathrm{sem}}^{k}
+
\lambda_{\mathrm{top}}\mathcal{L}_{\mathrm{top}}^{k}.
\label{eq:method_trace_loss}
\end{equation}
Specifically, $\mathcal{L}_{\mathrm{mod}}^{k}$ reconstructs modality-side information from the hierarchical trace state, $\mathcal{L}_{\mathrm{sem}}^{k}$ aligns modality evidence with the activated node-semantic trace, and $\mathcal{L}_{\mathrm{top}}^{k}$ encourages topology-context traces to preserve local structural relations. These terms keep trace paths faithful to modality contents, reusable semantic units, and topology context without exposing raw client data.

To stabilize trace units and avoid code collapse, FedLAB regularizes the typed codebooks:
\begin{equation}
\begin{aligned}
\mathcal{L}_{\mathrm{cb}}^{k}
=
\sum_{a}
\sum_{h\in\mathcal{H}_{k}^{a}}
\ell_q^a(h)
+
\eta
\sum_a
D_{\mathrm{KL}}
\left(
\bar{q}_{k}^{a}
\|
\operatorname{Unif}(L_a)
\right),
\end{aligned}
\label{eq:method_code_loss}
\end{equation}
where $\mathcal{H}_{k}^{a}$ is the set of local type-$a$ representations, $\bar{q}_{k}^{a}$ is the empirical usage distribution, and $\ell_q^a(h)$ is the standard quantization commitment loss. The usage regularizer prevents a small number of trace units from dominating the shared vocabulary.

\textbf{Barycenter refinement.}
After local training, client $k$ uploads model updates and aggregated posterior trace statistics. For type $a$ and code index $i$, FedLAB collects only posterior usage counts and posterior-weighted local centers:
\begin{equation}
n_{k,i}^{a}
=
\sum_{h\in\mathcal{H}_{k}^{a}}q_i^a(h),
\mu_{k,i}^{a}
=
\frac{
\sum_{h\in\mathcal{H}_{k}^{a}}q_i^a(h)h
}{
n_{k,i}^{a}+\epsilon
}.
\label{eq:method_stats}
\end{equation}
The server forms a usage-weighted semantic barycenter:
\begin{equation}
\hat{\mu}_{i}^{a,t}
=
\sum_k
\alpha_{k,i}^{a}\mu_{k,i}^{a},
\quad
\alpha_{k,i}^{a}
=
\frac{
n_{k,i}^{a}
}{
\sum_{k'}n_{k',i}^{a}+\epsilon
}.
\label{eq:method_barycenter}
\end{equation}
The global trace unit is then refined by
\begin{equation}
b_i^{a,t+1}
=
\operatorname{Proj}_{\mathbb{S}}
\left(
(1-\eta_c)b_i^{a,t}
+
\eta_c\hat{\mu}_{i}^{a,t}
\right).
\label{eq:method_code_agg}
\end{equation}
This update makes each global trace unit follow its cross-client posterior usage, aligning local modality, semantic, and topology traces into reusable foundation-level references.

The overall federated objective is
\begin{equation}
\mathcal{L}_{\mathrm{FedLAB}}
=
\sum_{k=1}^{K}
\frac{N_k}{N}
\mathcal{L}_{\mathrm{loc}}^{k},
\quad
N=\sum_{k=1}^{K}N_k .
\label{eq:method_final_loss}
\end{equation}
Thus, local trace preservation keeps trace paths faithful to private multimodal graphs, while semantic barycenter refinement makes the shared trace vocabulary reusable across clients without sharing raw data.

\section{Experiments}
\label{sec:experiments}

In this section, we provide a comprehensive empirical evaluation of FedLAB. We
begin by introducing the experimental setup, and then seek to answer the
following research questions: \textbf{Q1:} Does FedLAB achieve competitive
performance over strong federated multimodal graph learning baselines across
diverse benchmarks? \textbf{Q2:} What is the individual contribution of
modality evidence, node semantic, and topology context codebooks in FedLAB?
\textbf{Q3:} How robust is FedLAB under different hyper-parameter choices,
client conditions, and training dynamics? \textbf{Q4:} What computation and
memory efficiency does FedLAB achieve compared with existing baselines?
\textbf{Q5:} Can FedLAB provide faithful and reusable semantic traces for
multimodal federated graph predictions?

\subsection{Experimental Setup}
\label{sec:exp_setup}

\textbf{Datasets.}
We evaluate FedLAB on ten multimodal-attributed graph benchmarks~\cite{mmopenfgl} spanning
diverse domains: Toys, Grocery, Bili Music, DY, KU, Bili Food, QB,
Bili Cartoon, Flickr30k, and SemArt~\cite{ni2019_Grocery_Cloth_Ele_Movies_Sports,zhang2024ninerec,plummer2015_Flickr30k,garcia2018_SemArt}.
We construct federated scenarios by applying the Louvain community detection
algorithm~\cite{blondel2008louvain} to partition each benchmark into 10
non-IID clients.

\textbf{Baselines.}
We organize the compared methods into five groups. (1) \textbf{FL}:
FedAvg is used as the generic federated optimization baseline~\cite{fedavg}. (2)
\textbf{MM-GNN}: Fed-MGNet and Fed-MHGAT capture multimodal
graph dependencies with GNN-based architectures. (3) \textbf{MM-FL}:
FedMVP and FedMAC are representative multimodal federated
learning methods. (4) \textbf{GFM}: Fed-GFT and
Fed-GraphCLIP instantiate graph foundation model baselines. (5)
\textbf{FGL-GFM}: FedGFM+ and FedBook further introduce
foundation-model-style knowledge sharing in federated graph learning.
We use the corresponding representative sources for these baseline families:
MM-GNN~\cite{kong2021multiplex,jia2023multimodal},
MM-FL~\cite{che2024leveraging,nguyen2024fedmac},
GFM~\cite{wang2024gft,graphclip}, and FGL-GFM~\cite{fedgfm,fedbook}.

\textbf{Downstream Tasks.}
Following the multimodal federated graph benchmark protocol,
the evaluation covers two families of downstream tasks. The graph-centric
family consists of node classification and link prediction, where we use
Accuracy and AUC as metrics. The modality-centric family consists of modality
matching, modality retrieval, G2text, and G2image
generation, evaluated by AUC, Recall@5, ROUGE-L, and
CLIP-S, respectively. We
report mean test performance with standard deviation over repeated runs.

\subsection{Overall Performance}
\label{sec:overall_performance}

To answer Q1, we compare FedLAB with representative federated multimodal graph learning baselines across all downstream tasks. As shown in Table~\ref{table:mm_results}, FedLAB consistently achieves the best results on the evaluated benchmarks, with an average improvement of 4.38\% over the strongest competing baseline and a maximum gain of 7.53\%. The improvement is observed across both graph-centric tasks, such as node classification and link prediction, and modality-centric tasks, such as modality matching, retrieval, and generation. These results indicate that typed semantic codebooks do not merely improve a specific task type, but provide a reusable traceable representation that benefits heterogeneous multimodal graph learning under federated settings.

\providecommand{\res}{}
\renewcommand{\res}[2]{\mbox{$#1_{\scriptstyle \pm #2}$}}

\providecommand{\best}{}
\renewcommand{\best}[2]{\cellcolor[HTML]{DADADA}\mbox{$\mathbf{#1}_{\scriptstyle \pm #2}$}}

\providecommand{\second}{}
\renewcommand{\second}[2]{\underline{\res{#1}{#2}}}

\begin{table*}[t]
    \centering
    \caption{Overall performance on MM-FGL
    (mean $\pm$ std). Best results are \textbf{bold} and
    second-best are \underline{underlined}.}
    \label{table:mm_results}
    \setlength{\tabcolsep}{1.2pt}
    \renewcommand{\arraystretch}{1.18}
    \setlength{\extrarowheight}{0pt}

    \resizebox{\textwidth}{!}{
    \begin{tabular}{c|c|cc|cc|cc|cc|c|c}
    \specialrule{1.5pt}{1.5pt}{1.5pt}

    \multicolumn{2}{c|}{\textbf{Description}}
    & \multicolumn{2}{c|}{\begin{tabular}{@{}c@{}}\textbf{Node Classification}\\\textbf{(Acc)}\end{tabular}}
    & \multicolumn{2}{c|}{\begin{tabular}{@{}c@{}}\textbf{Link Prediction}\\\textbf{(AUC)}\end{tabular}}
    & \multicolumn{2}{c|}{\begin{tabular}{@{}c@{}}\textbf{Modality Matching}\\\textbf{(AUC)}\end{tabular}}
    & \multicolumn{2}{c|}{\begin{tabular}{@{}c@{}}\textbf{Modality Retrieval}\\\textbf{(Recall@5)}\end{tabular}}
    & \begin{tabular}{@{}c@{}}\textbf{G2Text}\\\textbf{(ROUGE-L)}\end{tabular}
    & \begin{tabular}{@{}c@{}}\textbf{G2Image}\\\textbf{(CLIP-S)}\end{tabular} \\

    \cmidrule(lr){1-2}
    \cmidrule(lr){3-4}
    \cmidrule(lr){5-6}
    \cmidrule(lr){7-8}
    \cmidrule(lr){9-10}
    \cmidrule(lr){11-11}
    \cmidrule(lr){12-12}

    \multicolumn{2}{c|}{\textbf{Method}}
    & \textbf{Toys}
    & \textbf{Grocery}
    & \textbf{Bili Music}
    & \textbf{DY}
    & \textbf{KU}
    & \textbf{Bili Food}
    & \textbf{QB}
    & \textbf{Bili Cartoon}
    & \textbf{Flickr30k}
    & \textbf{SemArt} \\
    \midrule

    \cellcolor[HTML]{EFD4D8}\textcolor{black}{\makebox[1.18cm][c]{\textbf{FL}}}
    & \cellcolor[HTML]{F8E9EC} FedAvg
    & \res{78.51}{0.09}
    & \res{80.10}{0.27}
    & \res{65.91}{0.18}
    & \res{65.15}{0.07}
    & \res{54.66}{0.53}
    & \res{54.63}{1.09}
    & \res{83.88}{1.74}
    & \res{73.51}{1.78}
    & \res{43.38}{0.32}
    & \res{70.12}{0.15} \\

    \cellcolor[HTML]{DCE8D7}
    & \cellcolor[HTML]{EEF5EB} Fed-MGNet
    & \res{67.02}{0.56}
    & \res{71.48}{1.02}
    & \res{68.15}{1.32}
    & \second{67.18}{0.25}
    & \res{57.61}{0.49}
    & \res{57.43}{1.00}
    & \res{84.38}{3.19}
    & \res{73.24}{2.49}
    & \res{48.34}{1.35}
    & \res{71.34}{0.37} \\

    \multirow{-2}{*}{\cellcolor[HTML]{DCE8D7}\textcolor{black}{\makebox[1.18cm][c]{\shortstack[c]{\textbf{MM}\\[-1pt]\textbf{GNN}}}}}
    & \cellcolor[HTML]{EEF5EB} Fed-MHGAT
    & \res{77.29}{1.57}
    & \res{78.06}{2.02}
    & \second{68.46}{2.01}
    & \res{63.80}{1.26}
    & \res{55.90}{1.39}
    & \res{57.90}{1.78}
    & \res{87.74}{4.89}
    & \res{71.47}{5.21}
    & \res{47.56}{1.18}
    & \res{57.40}{2.35} \\

    \cellcolor[HTML]{DCE8F2}
    & \cellcolor[HTML]{E3F2FD} FedMVP
    & \res{78.98}{0.07}
    & \res{79.93}{0.43}
    & \res{67.00}{0.47}
    & \res{64.75}{0.73}
    & \second{58.02}{0.47}
    & \second{58.97}{0.99}
    & \res{87.11}{4.97}
    & \res{70.36}{1.32}
    & \res{48.39}{1.24}
    & \res{70.19}{1.23} \\

    \multirow{-2}{*}{\cellcolor[HTML]{DCE8F2}\textcolor{black}{\makebox[1.18cm][c]{\shortstack[c]{\textbf{MM}\\[-1pt]\textbf{FL}}}}}
    & \cellcolor[HTML]{E3F2FD} FedMAC
    & \res{78.57}{0.25}
    & \res{80.06}{0.61}
    & \res{63.30}{0.51}
    & \res{64.51}{1.55}
    & \res{56.74}{0.37}
    & \res{57.40}{1.26}
    & \res{88.65}{2.75}
    & \res{75.15}{2.03}
    & \res{48.55}{1.31}
    & \res{71.37}{1.34} \\

    \cellcolor[HTML]{E8D8E4}
    & \cellcolor[HTML]{F5EDF3} Fed-GFT
    & \res{78.97}{0.31}
    & \second{80.30}{0.22}
    & \res{64.53}{0.19}
    & \res{63.68}{0.53}
    & \res{55.80}{0.80}
    & \res{57.67}{1.01}
    & \res{90.15}{1.80}
    & \res{71.95}{2.64}
    & \res{47.91}{1.17}
    & \res{72.03}{1.69} \\

    \multirow{-2}{*}{\cellcolor[HTML]{E8D8E4}\textcolor{black}{\makebox[1.18cm][c]{\textbf{GFM}}}}
    & \cellcolor[HTML]{F5EDF3} Fed-GraphCLIP
    & \res{76.34}{0.49}
    & \res{73.95}{0.35}
    & \res{66.68}{0.13}
    & \res{64.10}{0.06}
    & \res{55.63}{0.32}
    & \res{58.04}{1.27}
    & \second{91.23}{2.17}
    & \second{78.59}{5.33}
    & \res{47.51}{1.26}
    & \res{68.56}{0.12} \\

    \cellcolor[HTML]{EFEBCB}
    & \cellcolor[HTML]{FFFDE7} FedGFM+
    & \res{77.48}{0.36}
    & \res{73.81}{1.67}
    & \res{67.07}{0.12}
    & \res{63.54}{0.22}
    & \res{57.31}{0.37}
    & \res{58.23}{1.21}
    & \res{84.19}{1.65}
    & \res{76.26}{3.89}
    & \second{48.88}{1.22}
    & \res{73.27}{0.94} \\

    \multirow{-2}{*}{\cellcolor[HTML]{EFEBCB}\textcolor{black}{\makebox[1.18cm][c]{\shortstack[c]{\textbf{FGL}\\[-1pt]\textbf{GFM}}}}}
    & \cellcolor[HTML]{FFFDE7} FedBook
    & \second{80.19}{0.77}
    & \res{76.45}{0.17}
    & \res{66.29}{0.35}
    & \res{62.96}{0.28}
    & \res{57.63}{0.47}
    & \res{58.91}{0.28}
    & \res{77.44}{3.28}
    & \res{65.31}{2.57}
    & \res{46.57}{1.33}
    & \second{74.12}{2.68} \\
    \midrule

    \cellcolor[HTML]{DDF0D0}\textcolor{black}{\makebox[1.18cm][c]{\textbf{Ours}}}
    & \cellcolor[HTML]{E6FDD1} \textbf{FedLAB}
    & \best{84.14}{0.21}
    & \best{84.20}{0.15}
    & \best{74.32}{0.09}
    & \best{74.71}{0.07}
    & \best{63.50}{0.13}
    & \best{65.60}{0.93}
    & \best{94.49}{1.74}
    & \best{81.48}{1.75}
    & \best{49.33}{1.12}
    & \best{77.93}{0.39} \\

    \specialrule{1.3pt}{2.0pt}{1.0pt}
    \end{tabular}}
\end{table*}

\subsection{Ablation Study}
\label{sec:ablation}

To answer Q2, we remove each codebook from FedLAB while keeping the remaining
training protocol unchanged. As shown in Table~\ref{table:ablation}, removing
the topology, modality, and node codebooks reduces the average performance by
6.23\%, 6.91\%, and 7.81\%, respectively. The node codebook contributes most
to cross-client semantic reuse, the topology codebook supports structural
reasoning, and the modality codebook preserves fine-grained cross-modal
evidence, confirming that the three codebooks provide complementary signals.

\begin{table*}[t]
    \centering
    \caption{Ablation study on all datasets (mean $\pm$ std). Each variant
    removes one key codebook to validate its contribution.}
    \label{table:ablation}
    \setlength{\tabcolsep}{2.2pt}
    \renewcommand{\arraystretch}{1.08}
    \resizebox{\textwidth}{!}{
    \begin{tabular}{c|cc|cc|cc|cc|c|c}
    \specialrule{1.3pt}{1.2pt}{1.2pt}
    \multirow{2}{*}{\textbf{Variant}}
    & \multicolumn{2}{c|}{\textbf{Node Classification (Acc)}}
    & \multicolumn{2}{c|}{\textbf{Link Prediction (AUC)}}
    & \multicolumn{2}{c|}{\textbf{Modal Matching (AUC)}}
    & \multicolumn{2}{c|}{\textbf{Modal Retrieval (R@5)}}
    & \textbf{G2Text}
    & \textbf{G2Image} \\
    \cmidrule(lr){2-3}
    \cmidrule(lr){4-5}
    \cmidrule(lr){6-7}
    \cmidrule(lr){8-9}
    \cmidrule(lr){10-10}
    \cmidrule(lr){11-11}
    & \textbf{Toys} & \textbf{Grocery}
    & \textbf{Bili Music} & \textbf{DY}
    & \textbf{KU} & \textbf{Bili Food}
    & \textbf{QB} & \textbf{Bili Cartoon}
    & \textbf{Flickr30k}
    & \textbf{SemArt} \\
    \midrule
    Full FedLAB
    & \best{84.14}{0.21}
    & \best{84.20}{0.15}
    & \best{74.32}{0.09}
    & \best{74.71}{0.07}
    & \best{63.50}{0.13}
    & \best{65.60}{0.93}
    & \best{94.49}{1.74}
    & \best{81.48}{1.75}
    & \best{49.33}{1.12}
    & \best{77.93}{0.39} \\
    w/o Topology Codebook
    & \res{74.09}{0.31}
    & \second{77.36}{0.22}
    & \res{60.42}{0.36}
    & \res{60.65}{0.29}
    & \second{60.86}{0.24}
    & \res{62.37}{1.05}
    & \res{92.03}{1.03}
    & \res{79.68}{1.24}
    & \second{47.86}{1.09}
    & \second{72.10}{0.52} \\
    w/o Modality Codebook
    & \res{74.26}{0.27}
    & \res{76.89}{0.25}
    & \second{64.10}{0.33}
    & \second{61.37}{0.31}
    & \res{55.42}{0.31}
    & \res{60.94}{1.18}
    & \second{92.63}{0.96}
    & \second{80.37}{1.18}
    & \res{45.12}{1.21}
    & \res{69.52}{0.73} \\
    w/o Node Codebook
    & \second{76.48}{0.34}
    & \res{63.12}{0.41}
    & \res{64.07}{0.35}
    & \res{60.88}{0.30}
    & \res{57.18}{0.36}
    & \second{64.85}{1.11}
    & \res{91.24}{1.10}
    & \res{80.27}{1.21}
    & \res{44.76}{1.16}
    & \res{68.74}{0.68} \\
    \specialrule{1.3pt}{1.2pt}{1.2pt}
    \end{tabular}}
\end{table*}

\subsection{Robustness Analysis}
\label{sec:robustness}

To answer \textbf{Q3}, we evaluate FedLAB from three robustness perspectives:
hyper-parameter sensitivity, client scalability, and training dynamics.

\begin{figure}[!t]
    \centering
    \includegraphics[width=\linewidth,height=0.48\linewidth,keepaspectratio]{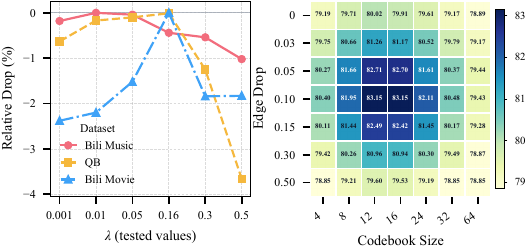}
    \caption{Robustness analysis under hyper-parameter variations. $\lambda$
    controls semantic code regularization, while edge-drop ratio and topology
    codebook size control topology perturbation strength and code granularity.}
    \label{fig:robust_hyper}
\end{figure}

\textbf{Hyper-parameter Sensitivity.}
We evaluate three key hyper-parameters: $\lambda$, edge-drop ratio, and
topology codebook size $L_t$, which control semantic code regularization,
topology perturbation strength, and topology-code granularity, respectively.
As shown in Fig.~\ref{fig:robust_hyper}, FedLAB remains stable across a broad
range of choices, and the heatmap presents a wide high-performing region,
suggesting that FedLAB does not rely on a brittle manually tuned configuration.

\begin{figure}[!t]
    \centering
    \includegraphics[width=\linewidth,height=0.48\linewidth,keepaspectratio]{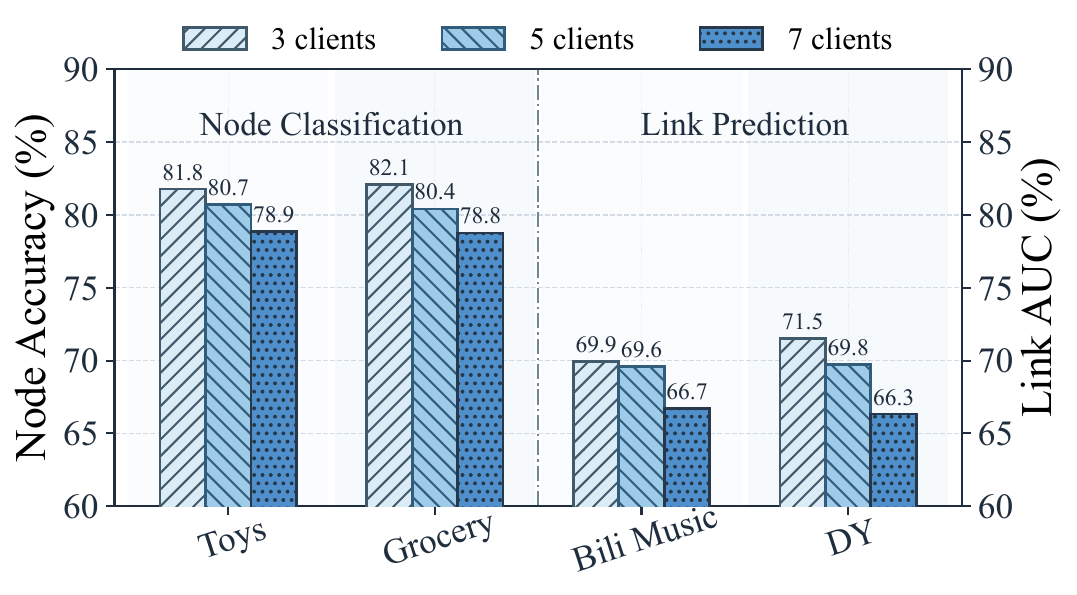}
    \caption{Robustness under different numbers of clients.}
    \label{fig:robust_client}
\end{figure}

\textbf{Client Scalability.}
We further evaluate FedLAB when federated partitions become more fragmented.
As shown in Fig.~\ref{fig:robust_client}, the performance changes mildly as
the number of clients increases, indicating that the shared traceable
codebooks help preserve transferable semantics under heterogeneous partitions.

\begin{figure}[t]
    \centering
    \includegraphics[width=\linewidth]{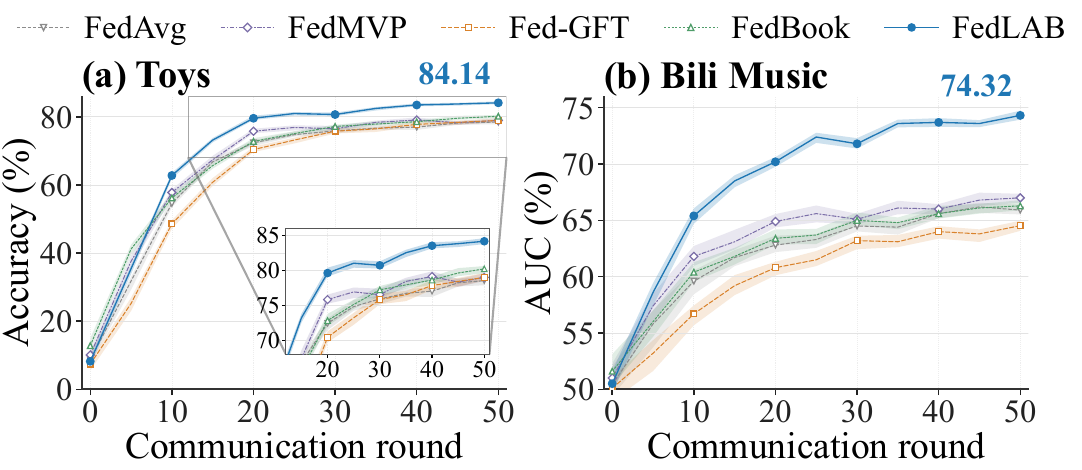}
    \caption{Convergence comparison across communication rounds on
    representative datasets.}
    \label{fig:convergence}
\end{figure}

\textbf{Training Dynamics.}
We finally compare convergence on graph-centric tasks. As shown in
Fig.~\ref{fig:convergence}, FedLAB converges smoothly and reaches better final
performance than baselines, showing that traceable codebook learning preserves
stable federated optimization.

\subsection{Efficiency Analysis}
\label{sec:efficiency}

\begin{figure}[t]
    \centering
    \includegraphics[width=\linewidth]{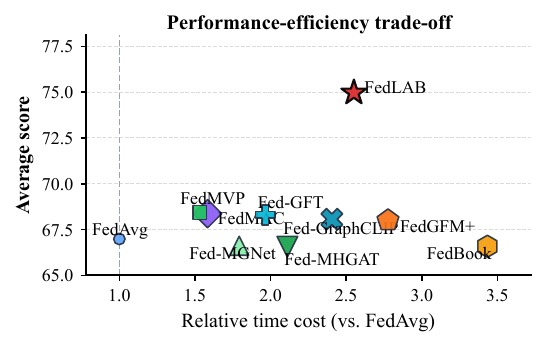}
    \caption{Performance-efficiency trade-off. The $x$-axis reports relative
    time cost against FedAvg, and the $y$-axis reports average performance
    across benchmarks.}
    \label{fig:eff_tradeoff}
\end{figure}

To answer Q4, we compare communication payload, computation, space cost, and
wall-clock time. As shown in Table~\ref{table:efficiency}, FedLAB requires
additional communication and memory for traceable codebooks, but remains faster
than FedBook. Fig.~\ref{fig:eff_tradeoff} further shows that FedLAB achieves
the highest average performance with a practical time cost.

\begin{table}[t]
    \centering
    \caption{Efficiency and cost analysis. FedLAB incurs additional trace-codebook
overhead for semantic traceability, while maintaining practical wall-clock
time compared with foundation-style baselines.}
    \label{table:efficiency}
    \setlength{\tabcolsep}{3.0pt}
    \renewcommand{\arraystretch}{1.12}
    \resizebox{\linewidth}{!}{
    \begin{tabular}{c|c|c|c|c}
    \specialrule{1.2pt}{1.0pt}{1.0pt}
    \textbf{Method} &
    \begin{tabular}{@{}c@{}}\textbf{Comm. Payload}\\\textbf{(Scalars)}\end{tabular} &
    \begin{tabular}{@{}c@{}}\textbf{Total Ops}\\\textbf{(FLOPs)}\end{tabular} &
    \begin{tabular}{@{}c@{}}\textbf{Space Cost}\\\textbf{(MB)}\end{tabular} &
    \begin{tabular}{@{}c@{}}\textbf{Time}\\\textbf{(s)}\end{tabular} \\
    \midrule
    FedAvg & $2.96\times10^{6}$ & $1.47\times10^{9}$ & 63.9 & 74.09 \\
    FedMVP & $7.72\times10^{6}$ & $1.49\times10^{9}$ & 507.7 & 117.33 \\
    FedMAC & $3.96\times10^{5}$ & $1.69\times10^{9}$ & 83.2 & 113.44 \\
    FedBook & $3.07\times10^{8}$ & $2.33\times10^{10}$ & 844.2 & 254.39 \\
    \rowcolor[HTML]{DADADA}
    FedLAB & $6.66\times10^{8}$ & $2.40\times10^{10}$ & 1583.3 & 188.99 \\
    \specialrule{1.2pt}{1.0pt}{1.0pt}
    \end{tabular}}
\end{table}

\subsection{Semantic Traceability Validation}
\label{sec:explainability}

To answer \textbf{Q5}, we assess whether FedLAB produces prediction-supporting
and reusable traces for modality evidence, topology context, and node semantics.

\begin{figure}[t]
    \vspace{-8pt}
    \centering
    \includegraphics[width=\linewidth,trim=0 14pt 0 0,clip]{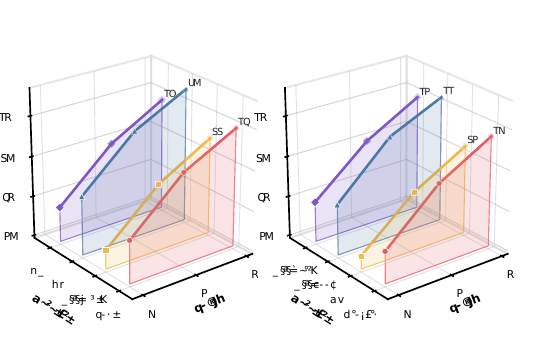}
    \vspace{-2pt}
    \caption{Modality evidence traceability.}
    \label{fig:exp_modality_sufficiency}
\end{figure}

\textbf{Modality Evidence Traceability.}
For instance $v$, let $E_v^K$ denote the top-$K$ traced modality evidence codes
and $X_v$ denote the full multimodal input. We define trace sufficiency as
\begin{equation}
    \mathrm{TS}@K=\mathbb{E}_{v}
    \frac{p(y_v\mid E_v^K)}{p(y_v\mid X_v)} ,
    \label{eq:trace_sufficiency}
\end{equation}
which measures the prediction confidence retained by selected trace units. As
shown in Fig.~\ref{fig:exp_modality_sufficiency}, FedLAB preserves more
confidence as $K$ increases, indicating that its modality codes capture compact
prediction-supporting signals.

\begin{figure}[!t]
    \vspace{-2pt}
    \centering
    \includegraphics[width=\linewidth]{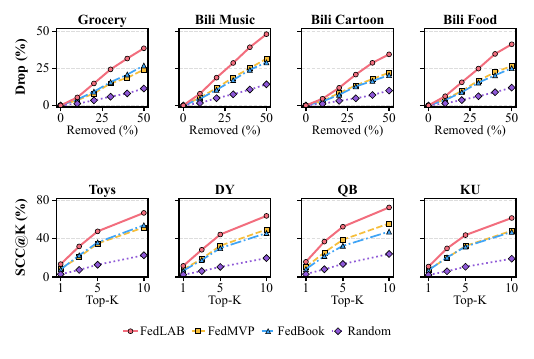}
    \vspace{-2pt}
    \caption{Semantic traceability validation on topology trace drop (upper) and
semantic code concentration (lower).}
    \label{fig:exp_explainability_compact}
\end{figure}

\textbf{Topology Context Traceability.}
We next evaluate whether topology traces capture structural support behind
predictions. Let $\mathcal{G}_v$ denote the local topology and
$\mathcal{R}_v^r$ denote the top-$r$ percent removed topology trace units. We
measure topology trace drop as
\begin{equation}
    \mathrm{TTD}(r)=\mathbb{E}_{v}
    \left[
    p(y_v\mid\mathcal{G}_v)-
    p(y_v\mid\mathcal{G}_v\setminus\mathcal{R}_v^r)
    \right],
    \label{eq:topology_trace_drop}
\end{equation}
where a larger value indicates stronger structural decision support. As shown
in the upper panels of Fig.~\ref{fig:exp_explainability_compact}, removing
FedLAB-selected topology traces causes steeper confidence drops than post-hoc
or random removal baselines~\cite{gnnexplainer,pgexplainer,graphlime}.

\textbf{Reusable Node Semantic Codes.}
Finally, we test whether node-semantic codes form reusable references. For
label $c$, let $\mathcal{V}_c$ denote its nodes and $\mathcal{T}_c^K$ be the
top-$K$ most frequent semantic codes in $\mathcal{V}_c$. We define semantic code
concentration as
\begin{equation}
    \mathrm{SCC}@K
    =
    \frac{1}{|\mathcal{Y}|}
    \sum_{c\in\mathcal{Y}}
    \frac{1}{|\mathcal{V}_c|}
    \sum_{v\in\mathcal{V}_c}
    \mathbb{I}[c_v^s\in\mathcal{T}_c^K],
    \label{eq:scc}
\end{equation}
where $c_v^s$ is the node-semantic code assigned to $v$. As shown in the lower
panels of Fig.~\ref{fig:exp_explainability_compact}, FedLAB achieves higher
SCC@K than non-traceable and random assignment baselines, suggesting that
same-label instances are covered by compact reusable codes rather than scattered
local representations.

\section{Conclusion}
\label{sec:conclusion}

In this paper, we study Fed-MMGFM from the
perspective of semantic traceability. We identify that existing federated graph
learning and graph foundation models can transfer useful representations, but
their predictions remain difficult to audit because modality evidence, node
semantics, and topology context are entangled in opaque embeddings. Motivated
by this observation, we propose FedLAB, a traceable semantic codebook framework
that organizes federated multimodal graph knowledge into typed modality
evidence, node semantic, and topology context codebooks. FedLAB constructs
shared semantic references from decentralized multimodal graphs and refines
them through federated foundation pre-training while keeping raw modalities and
local graph structures private. Extensive experiments on ten benchmarks and
six downstream tasks demonstrate that FedLAB consistently improves overall
performance, preserves robustness under different client and hyper-parameter
conditions, and maintains practical efficiency. Further explainability
validation shows that FedLAB provides faithful and reusable semantic traces,
offering a promising foundation for auditable multimodal graph intelligence in
federated environments.

\bibliographystyle{aaai}
\bibliography{fedlab_refs}

@inproceedings{fedavg,
  title={Communication-efficient learning of deep networks from decentralized data},
  author={McMahan, Brendan and Moore, Eider and Ramage, Daniel and Hampson, Seth and y Arcas, Blaise Aguera},
  booktitle={Artificial intelligence and statistics},
  pages={1273--1282},
  year={2017},
  organization={Pmlr}
}

@article{openfgl,
  title={Openfgl: A comprehensive benchmark for federated graph learning},
  author={Li, Xunkai and Zhu, Yinlin and Pang, Boyang and Yan, Guochen and Yan, Yeyu and Li, Zening and Wu, Zhengyu and Zhang, Wentao and Li, Rong-Hua and Wang, Guoren},
  journal={arXiv preprint arXiv:2408.16288},
  year={2024}
}

@article{mmopenfgl,
  title={MM-OpenFGL: A Comprehensive Benchmark for Multimodal Federated Graph Learning},
  author={Li, Xunkai and Ai, Yuming and Zhu, Yinlin and Lu, Haodong and Zhang, Yi and Fu, Guohao and Fan, Bowen and Dai, Qiangqiang and Li, Rong-Hua and Wang, Guoren},
  journal={arXiv preprint arXiv:2601.22416},
  year={2026}
}

@inproceedings{graphmae,
  title={Graphmae: Self-supervised masked graph autoencoders},
  author={Hou, Zhenyu and Liu, Xiao and Cen, Yukuo and Dong, Yuxiao and Yang, Hongxia and Wang, Chunjie and Tang, Jie},
  booktitle={Proceedings of the 28th ACM SIGKDD conference on knowledge discovery and data mining},
  pages={594--604},
  year={2022}
}

@article{fedgfm,
  title={Towards effective federated graph foundation model via mitigating knowledge entanglement},
  author={Zhu, Yinlin and Li, Xunkai and Jia, Jishuo and Hu, Miao and Wu, Di and Qiu, Meikang},
  journal={Advances in Neural Information Processing Systems},
  volume={38},
  pages={64599--64628},
  year={2026}
}

@article{fedbook,
  title={FedBook: A Unified Federated Graph Foundation Codebook with Intra-domain and Inter-domain Knowledge Modeling},
  author={Wu, Zhengyu and Zhu, Yinlin and Li, Xunkai and Qiu, Ziang and Li, Rong-Hua and Wang, Guoren and Zhou, Chenghu},
  journal={arXiv preprint arXiv:2510.07755},
  year={2025}
}

@inproceedings{graphclip,
  title={Graphclip: Enhancing transferability in graph foundation models for text-attributed graphs},
  author={Zhu, Yun and Shi, Haizhou and Wang, Xiaotang and Liu, Yongchao and Wang, Yaoke and Peng, Boci and Hong, Chuntao and Tang, Siliang},
  booktitle={Proceedings of the ACM on Web Conference 2025},
  pages={2183--2197},
  year={2025}
}

@inproceedings{unigraph2,
  title={Unigraph2: Learning a unified embedding space to bind multimodal graphs},
  author={He, Yufei and Sui, Yuan and He, Xiaoxin and Liu, Yue and Sun, Yifei and Hooi, Bryan},
  booktitle={Proceedings of the ACM on Web Conference 2025},
  pages={1759--1770},
  year={2025}
}

@article{gnnexplainer,
  title={Gnnexplainer: Generating explanations for graph neural networks},
  author={Ying, Zhitao and Bourgeois, Dylan and You, Jiaxuan and Zitnik, Marinka and Leskovec, Jure},
  journal={Advances in neural information processing systems},
  volume={32},
  year={2019}
}

@article{pgexplainer,
  title={Parameterized explainer for graph neural network},
  author={Luo, Dongsheng and Cheng, Wei and Xu, Dongkuan and Yu, Wenchao and Zong, Bo and Chen, Haifeng and Zhang, Xiang},
  journal={Advances in neural information processing systems},
  volume={33},
  pages={19620--19631},
  year={2020}
}

@article{graphlime,
  title={Graphlime: Local interpretable model explanations for graph neural networks},
  author={Huang, Qiang and Yamada, Makoto and Tian, Yuan and Singh, Dinesh and Chang, Yi},
  journal={IEEE Transactions on Knowledge and Data Engineering},
  volume={35},
  number={7},
  pages={6968--6972},
  year={2022},
  publisher={IEEE}
}

@inproceedings{ni2019_Grocery_Cloth_Ele_Movies_Sports,
  title={Justifying recommendations using distantly-labeled reviews and fine-grained aspects},
  author={Ni, Jianmo and Li, Jiacheng and McAuley, Julian},
  booktitle={Proceedings of the Conference on Empirical Methods in Natural Language Processing and the International Joint Conference on Natural Language Processing, EMNLP-IJCNLP},
  year={2019}
}

@inproceedings{garcia2018_SemArt,
  title={How to read paintings: semantic art understanding with multi-modal retrieval},
  author={Garcia, Noa and Vogiatzis, George},
  booktitle={Proceedings of the European Conference on Computer Vision Workshops, ECCV},
  year={2018}
}

@inproceedings{plummer2015_Flickr30k,
  title={Flickr30k entities: Collecting region-to-phrase correspondences for richer image-to-sentence models},
  author={Plummer, Bryan A and Wang, Liwei and Cervantes, Chris M and Caicedo, Juan C and Hockenmaier, Julia and Lazebnik, Svetlana},
  booktitle={Proceedings of the IEEE International Conference on Computer Vision, ICCV},
  year={2015}
}

@article{blondel2008louvain,
  title={Fast unfolding of communities in large networks},
  author={Blondel, Vincent D and Guillaume, Jean-Loup and Lambiotte, Renaud and Lefebvre, Etienne},
  journal={Journal of statistical mechanics: theory and experiment},
  volume={2008},
  number={10},
  pages={P10008},
  year={2008},
  publisher={IOP Publishing}
}

@article{van2017neural,
  title={Neural discrete representation learning},
  author={Van Den Oord, Aaron and Vinyals, Oriol and others},
  journal={Advances in neural information processing systems},
  volume={30},
  year={2017}
}

@inproceedings{wang2025learning,
  title={Learning graph quantized tokenizers},
  author={Wang, Limei and Hassani, Kaveh and Zhang, Si and Fu, Dongqi and Yuan, Baichuan and Cong, Weilin and Hua, Zhigang and Wu, Hao and Yao, Ning and Long, Bo},
  booktitle={International Conference on Learning Representations},
  volume={2025},
  pages={97239--97260},
  year={2025}
}

@article{kong2021multiplex,
  title={Multiplex graph networks for multimodal brain network analysis},
  author={Kong, Zhaoming and Sun, Lichao and Peng, Hao and Zhan, Liang and Chen, Yong and He, Lifang},
  journal={arXiv preprint arXiv:2108.00158},
  year={2021}
}

@article{jia2023multimodal,
  title={Multimodal heterogeneous graph attention network},
  author={Jia, Xiangen and Jiang, Min and Dong, Yihong and Zhu, Feng and Lin, Haocai and Xin, Yu and Chen, Huahui},
  journal={Neural Computing and Applications},
  volume={35},
  number={4},
  pages={3357--3372},
  year={2023},
  publisher={Springer}
}

@inproceedings{che2024leveraging,
  title={Leveraging foundation models for multi-modal federated learning with incomplete modality},
  author={Che, Liwei and Wang, Jiaqi and Liu, Xinyue and Ma, Fenglong},
  booktitle={Joint European Conference on Machine Learning and Knowledge Discovery in Databases},
  pages={401--417},
  year={2024},
  organization={Springer}
}

@inproceedings{nguyen2024fedmac,
  title={Fedmac: Tackling partial-modality missing in federated learning with cross-modal aggregation and contrastive regularization},
  author={Nguyen, Manh Duong and Nguyen, Trung Thanh and Pham, Huy Hieu and Hoang, Trong Nghia and Le Nguyen, Phi and Huynh, Thanh Trung},
  booktitle={2024 22nd International Symposium on Network Computing and Applications (NCA)},
  pages={278--285},
  year={2024},
  organization={IEEE}
}

@article{wang2024gft,
  title={Gft: Graph foundation model with transferable tree vocabulary},
  author={Wang, Zehong and Zhang, Zheyuan and Chawla, Nitesh V and Zhang, Chuxu and Ye, Yanfang},
  journal={Advances in neural information processing systems},
  volume={37},
  pages={107403--107443},
  year={2024}
}

@article{zhang2024ninerec,
  title={Ninerec: A benchmark dataset suite for evaluating transferable recommendation},
  author={Zhang, Jiaqi and Cheng, Yu and Ni, Yongxin and Pan, Yunzhu and Yuan, Zheng and Fu, Junchen and Li, Youhua and Wang, Jie and Yuan, Fajie},
  journal={IEEE Transactions on Pattern Analysis and Machine Intelligence},
  year={2024},
  publisher={IEEE}
}

@article{chen2026stage,
  title={STAGE: Tackling Semantic Drift in Multimodal Federated Graph Learning},
  author={Chen, Zekai and Wu, Xun and Li, Xunkai and Sun, Yihan and Li, Rong-Hua and Wang, Guoren},
  journal={arXiv preprint arXiv:2605.11919},
  year={2026}
}

@article{chen2026prism,
  title={PRISM: Topology-Aware Cross-Modal Imputation for Modality-Deficient Federated Graph Learning},
  author={Chen, Zekai and Zhang, Miao and Xing, Jiayang and Li, Xunkai and Wu, Xun and Li, Rong-Hua and Wang, Guoren},
  journal={arXiv preprint arXiv:2606.09301},
  year={2026}
}

\end{document}